%% file: emnlp2020.tex
\documentclass[11pt,a4paper]{article}
\usepackage[hyperref]{emnlp2020}
\usepackage{times}
\usepackage{latexsym}

\usepackage{microtype}

\aclfinalcopy %

\newcommand{\pheadNoSpace}[1] {\noindent\textbf{#1.}} %
\newcommand{\pheadWithSpace}[1] {\vspace{1.25mm}\noindent\textbf{#1.}} %

\usepackage{booktabs} %
\usepackage{latexsym}
\usepackage{makecell}
\usepackage{booktabs} %
\usepackage{graphicx}
\graphicspath{ {images/} }
\usepackage{scalerel} %
\usepackage{multirow} %
\usepackage{makecell}
\usepackage{amsmath}
\usepackage{tcolorbox}
\usepackage{CJKutf8}
\usepackage{enumitem}
\usepackage{color}
\usepackage{algorithm}
\usepackage{algorithmic}
\usepackage{xcolor,colortbl}
\usepackage{array}%
\usepackage{multirow}

\definecolor{mygreen}{RGB}{34, 139, 34}

\title{Are Emojis Emotional?\\A Study to Understand the Association between Emojis and Emotions}

\author{Abu Awal Md Shoeb  \\
  Dept. of Computer Science \\
  Rutgers University \\
  New Brunswick, NJ, USA \\
  \texttt{abu.shoeb@rutgers.edu} \\\And
  Gerard de Melo \\
  Dept. of Computer Science \\
  Rutgers University \\
  New Brunswick, NJ, USA \\
  \texttt{gerard.demelo@rutgers.edu} \\}

\date{}

\begin{document}
\maketitle
\begin{abstract}
Given the growing ubiquity of emojis in language, there is a need for methods and resources that shed light on their meaning and communicative role. One conspicuous aspect of emojis is their use to convey affect in ways that may otherwise be non-trivial to achieve. In this paper, we seek to explore the connection between emojis and emotions by means of a new dataset consisting of human-solicited association ratings.
We additionally conduct experiments to assess to what extent such associations can be inferred from existing data, such that similar associations can be predicted for a larger set of emojis.
Our experiments show that this succeeds when high-quality word-level information is available.
\end{abstract}

\input{sec-intro.tex}

\input{sec-background.tex}
\input{sec-annotation-task.tex}

\input{sec-prediction.tex}
\input{sec-conclusion.tex}

\section*{Acknowledgments}

We sincerely thank all our annotators who volunteered to contribute to our study and helped us to establish this new resource for the community. Clearly, this research would not have been possible without their time and effort.


\input{references.bbl}
\end{document}

%% file: sec-intro.tex
\section{Introduction}
\label{sec:intro}

People increasingly rely on digital channels such as mobile instant messaging apps to communicate with their friends, families, colleagues, and communities.
Along with this rapid shift in medium, there have been concomitant changes in the way people express themselves in written language \cite{McCulloch2019BecauseInternet}.

One notable development has been the emergence of emojis as a novel, more visual form of expression. As a new modality, emojis present rich  possibilities for representation and interaction.
Emojis have become ubiquitous in social media and in instant messaging, owing in part to their visual appeal and their ease of use compared to typing out full words using virtual keyboards on mobile devices. 

However, the rise of emojis also substantially appears to stem from their ability to convey affect \cite{Vidal2016}.
This is evinced by the fact that the most frequently used emojis are smileys and other facial expression symbols that exhibit a direct connection to emotional expression.
This is further corroborated by the fact that traditional \emph{emoticons} such as ``:-)'' and ``:)'', now largely displaced by emojis, were chiefly used to convey humor and emotion, as also reflected in their name, a portmanteau of  \emph{emotion} and \emph{icon}.

This mandates additional analysis of the nexus between emojis and emotion. Past work has already compiled a list of sentiment polarity scores for a set of emojis \cite{novak2015}. Another study categorized a small set of 15 emojis into 4 different emotion classes \cite{emoji2emotion}. Further studies explored the linguistic connection between words and emojis \cite{Cappallo2019,barbieri-etal-2017-emojis,Naaman2017-mojisem,ShoebRajiDeMelo2019EmoTag}. However, none of the previous works assessed to what extent different emojis convey different emotions. 

In this work, we first conduct an annotation experiment to obtain real-valued emotion intensity scores for a set of emojis. Specifically, we solicit emotion intensity scores for 150 most popular emojis on Twitter according to Emoji Tracker\footnote{\url{http://emojitracker.com/}}, a platform that captures the real-time use of emojis on Twitter, obtaining data from 9 human raters.
The purpose of this endeavor is to measure the degree of emotion that a user associates with the use of a given emoji in written expression.
The resulting intensity scores are collected for eight different emotions, viz.\ \emph{anger}, \emph{anticipation}, \emph{disgust}, \emph{fear}, \emph{joy}, \emph{sadness}, \emph{surprise}, and \emph{trust}. 

We assess the emotion scores of this group of emojis and subsequently study a series of simple unsupervised models to predict such emotion intensity scores automatically. In order to support the prediction model, we consider an emoji-centric corpus consisting of 20M tweets and study how it can expose semantic relationships between emotion words and emojis, drawing on additional lexical resources.
The success of these scoring models enables us to also predict emotion intensity scores for further emojis not in our human-compiled data, which we will as well release.

%% file: sec-background.tex
\section{Background and Related Work}
\label{sec:background}

\pheadNoSpace{Emotion and Communication}
\newcite{Darwin1872} was among the first to consider the connection between emotions and their expression in substantial detail. He remarked for instance, that for both animals and humans, anger coincides with eye muscle contractions and teeth exposure, and commented on the fact that humans lift their eyebrows in moments of surprise. His work then goes on to study the role of such forms of facial expression in conveying to others how an animal feels, studying primates as well as human infants and adults. 

In light of this, humans continue to rely extensively on such nonverbal cues even in oral forms of linguistic communication.
Although a person's emotion and mood can to some extent be conveyed by means of suitable content words (e.g., ``I am happy to hear that!'') or interjections (``Wow!''), 
face-to-face communication has important properties that written communication tends to lack \cite{Bordia2017}. These include facial expressions of the aforementioned sort, but also gesture and intonation. In certain circumstances, e.g., certain problem-solving settings, face-to-face communication may hence prove more efficient and effective \cite{Bordia2017}.

Accordingly, since the beginning of writing, humans have resorted to surrogate mechanisms to convey emotive signals, attempting to push the boundaries and overcome some of the inherent restrictions of plain written language as a medium, e.g., by means of illustrative embellishments and ornaments.

\pheadWithSpace{Emoticons}
Emoticons such as ``:-)'' and Japanese \mbox{\begin{CJK}{UTF8}{min}顔文字\end{CJK}} (\emph{kaomoji}) such as ``(\textasciicircum\_\textasciicircum)'',
both based on regular characters, have been in use for several decades.
\newcite{Go_Bhayani_Huang_2009} proposed a form of distant supervision by using emoticons as noisy labels for Twitter sentiment classification. 
\newcite{Davidov:2010:ESL} adopted a similar approach by handpicking smileys and hashtags as tweet labels to train a supervised model to classify the sentiment of tweets.

\pheadWithSpace{Emojis} Emoji characters, similar to earlier dingbat characters, are pictorial but also colorful. Despite the lexicographic similarity between the two words \emph{emoji} and \emph{emotion}, etymo\-log\-ic\-al\-ly, the former stems from the Japanese words \begin{CJK}{UTF8}{min}絵\end{CJK} (\emph{e}, picture) and \mbox{\begin{CJK}{UTF8}{min}文字\end{CJK}} (\emph{moji}, character).

Emojis originated in Japan in the 1990s and have only recently spread globally. Historically, the spread of emojis has been driven in large part by their adoption in popular messaging and social media platforms, which led, among things, to their inclusion in Shift JIS, and, subsequently, the Unicode standard. Nowadays, they are ubiquitous in social media and chat applications, but increasingly also in emails and other digital correspondence.

Emojis have a number of different roles. 
\newcite{Kaye2017} explained how emojis may aid the interlocutor in disambiguating utterances that would otherwise remain ambiguous. 

One of their principal uses has been to convey emotion, particularly via facial expression emojis, as explained in Section \ref{sec:intro}. In 2015, Oxford Dictionaries declared the \emph{Face with Tears of Joy} emoji its Word of the Year 2015. Emojis may also be useful as a more instantaneously and widely recognized form of communicating degrees of satisfaction. \newcite{Kaye2017} go as far as suggesting them for consideration as possible alternatives to regular Likert scales.

\pheadWithSpace{Emoji Semantics} 
The MIT DeepMoji project \cite{deepmoji} developed a model that recommends emojis given a natural language sentence as input. A deep neural architecture was trained on a collection of 1.2B tweets to learn the sentiment, emotions, and the use of sarcasm in short text.

\newcite{emoji2016LREC} proposed a method to learn vector space embeddings of emojis using the standard word2vec skip-gram approach, applied to a large collection of tweets.
In contrast, 
\newcite{Eisner-2016} attempted to learn vector embeddings of emojis based on their short descriptions in the Unicode standard.
EmojiNet \cite{EmojiNet} provides a sense inventory to distinguish different senses of an emoji,
drawing on Web-crawled emoji definitions and connecting them to word senses from a lexical resource, along with vector representations of context words.

The first paper that thoroughly investigated the sentiment of emojis \cite{novak2015} proposed a sentiment ranking of 715 emojis on a corpus of 70,000 tweets. This work provides a basis for future research on the logographic usage of emojis in social media.

\newcite{emoji2emotion} proposed a method to classify emojis with regard to their sentiment and emotion. Their corpus consists of 500 labeled tweets,
and they categorize 15 emojis by classifying them with respect to 4 emotions.
For this, they applied a distant supervision technique for a reliable  mapping based on manually annotated data.

\newcite{mojitalk} trained a natural language conversation model that accounts for the underlying emotion of utterances by exploiting the existence of emojis as a signal.

%% file: sec-annotation-task.tex
\section{Annotation Task}
\label{sec:annotation-task}

In order to better study the connection between emojis and emotions, we proceeded to compile a dataset of ratings quantifying the perceived strength of association between emoijs and emotions.

\subsection{Task Setup and Guidelines}
\label{subsec:experiment-setup}

\paragraph{Target Emoji Set.} 
We considered a set of 150 most frequently used emojis, based on frequencies reported by the Emoji Tracker service\footnote{\url{http://emojitracker.com/}}, a platform that visualizes the real-time use of emojis on Twitter. The counters on Emoji Tracker indicate how many times an emoji has been used on Twitter since July 4, 2013. We rank all emojis based on their reported total frequency counts as of July 3, 2019 and pick the top 150 emojis for our annotation task. 

\paragraph{Emotion Set.} 
While numerous emotion models and affective classification schemes have been proposed, for this study we consider the 8 basic emotions proposed in the widely known Plutchik Wheel of Emotions model \cite{Plutchik1980}, i.e., \emph{anger}, \emph{anticipation}, \emph{disgust}, \emph{fear}, \emph{joy}, \emph{sadness}, \emph{surprise}, and \emph{trust}. 

\paragraph{Ratings.}
For a given emoji, the participants were asked to assess to what extent said emoji is associated with a given emotion. Note that association is a broad notion that not only covers emojis that are directly invoked to express an emotion, as in the case of certain facial expression emojis, but also encompasses mere conceptual association. For instance, the wrapped gift emoji may be associated with \emph{joy}, although the semantics of the emoji itself correspond to a \emph{present} or \emph{gift} rather than directly conveying \emph{joy}. This notion of association reflects a general, abstract form of connection, much like a prior. 
Clearly, embedded in a specific utterance, the specific emotions that are evoked may differ quite substantially, due to the complex ways in which different words along with embedded emojis interact to give rise to an overall interpretation. In this regard, our ratings are similar to lexical resources that seek to quantify context-independent lexical associations \cite{Hill2015SimLex} or word--emotion associations \cite{Mohammad13,Saif-LREC18}.

The degree of association was specified numerically as a score ranging from 0 (no association with the emotion) to  4 (representing the highest degree of association with the emotion). 
While we are cognizant of the challenges of directly eliciting scalar ratings from the annotators, we opted to follow prominent previous work on collecting association ratings \cite{Rubenstein:1965:CCS:365628.365657,Finkelstein:2001:PSC:371920.372094,Hill2015SimLex,gerz-etal-2016-simverb} in order to make our data comparable to such efforts.

\subsection{Data Collection}
\label{subsec:guidelines}

\paragraph{Interface.}
We developed a web interface to collect ratings.
We randomly split the target set of 150 emojis into a total of 6 subsets, each consisting of 25 emojis. 
When a rater selects a set from the main page, the corresponding 25 emojis load into a single page covering all 8 emotions. 

Within each set, we randomize the order of displayed emojis upon each page load, such
that different raters do not observe and annotate them in the same order. This ensures that different emojis within a set are given equal attention on average when aggregating scores from different human raters, mitigating potential fatigue-driven biases in the final ratings.

In total, an annotator needs to make 8 selections for a single emoji, corresponding to the set of 8 basic emotions. We ask users to provide all 8 emotions ratings for every single emoji. 
For each pairing of emoji and emotion, the rater needs to select the respective intensity level (from 0 to 4), provided as radio buttons on the page. 

\paragraph{Participants.}
We asked a total of 9 different human participants to each rate 150 emojis for 8 different emotions. All selected participants were native or near-native speakers of English and reported having extensive prior familiarity with emojis in their personal communication or from social media use. As mentioned, the emojis were grouped into 6 sets, each consisting of 25 emojis. The annotators were asked to annotate one such set per day so as to avoid overburdening them, which might
affect the quality of the rating.

Ultimately, for each pairing of emoji and emotion, we treat the mean value across the 9 individual raters as a real-valued score in $[0,4]$ reflecting the association for that pairing. We also compute for each pairing the standard deviation among its ratings.

\begin{figure}
    \centering
    \includegraphics[width=0.95\linewidth]{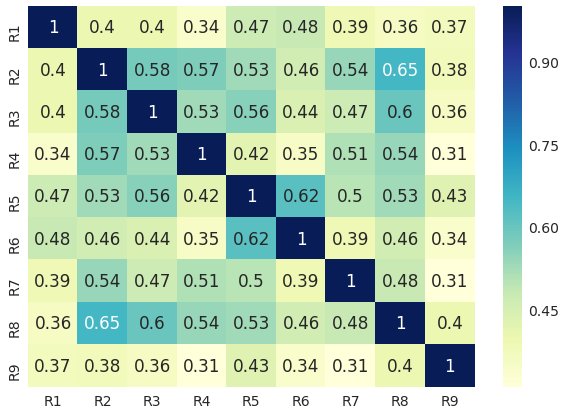}
    \caption{Pairwise Pearson Correlation for 9 raters based on all 8 emotion scores for the set of 150 most popular Twitter emojis}
    \label{fig:pearson-all-raters}
\end{figure}

\input{tab-score-details.tex}

\input{tab-emoji-group.tex}

\subsection{Analysis}
\label{subsec:inter-annotator-agreement}

In total, we collect 10,800 ratings for 1,200 pairings of emoji and emotion, covering 150 emojis, each rated with regard to 8 emotions by 9 human raters.

\paragraph{Inter-Annotator-Agreement.} To evaluate the agreement between the raters, 
we first check the overall agreement between pairs of human raters across the entire set of emoji--emotion ratings. 
This was in part also motivated by quality control concerns, i.e., a desire to assess whether there was any individual rater that disagreed substantially with all other raters.
Fortunately, this was not the case and we decided not to eliminate data from any rater.
Figure \ref{fig:pearson-all-raters} reports the pairwise Pearson Correlation scores between raters.

\begin{figure}
    \centering
    \includegraphics[width=0.95\linewidth]{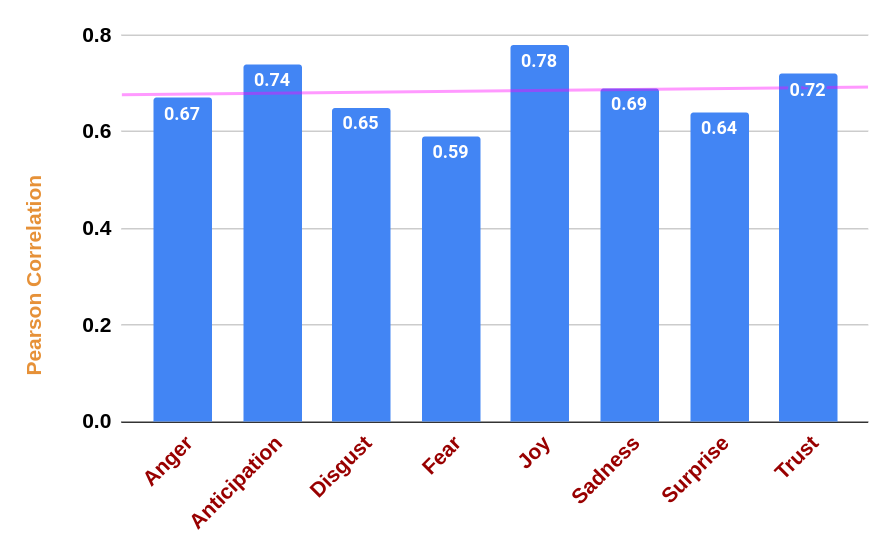}
    \caption{Average Pearson correlation coefficient between raters score and the gold score grouped by emotions. Pink line represents the overall trend}
    \label{fig:pearson-by-emotion}
\end{figure}

\begin{figure}
    \centering
    \includegraphics[width=0.90\linewidth]{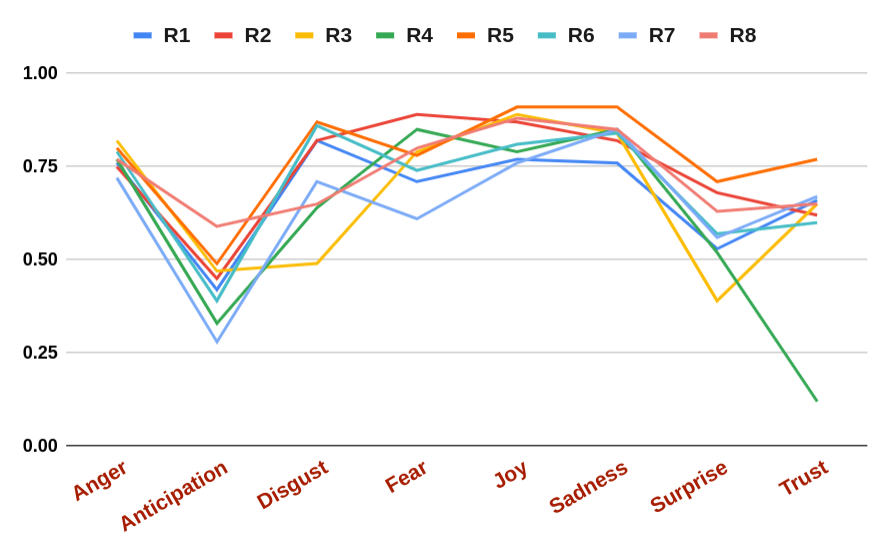}
    \caption{Variation of agreement for individual raters across emotions}
    \label{fig:pearson-fluctuation}
\end{figure}

In Figure \ref{fig:pearson-by-emotion}, we consider separately for different emotions the average agreement (Pearson correlation) of raters with the mean ratings. 
A fairly high agreement is observed for \emph{sadness}, \emph{joy}, and \emph{fear}. In contrast, 
we conjecture that for \emph{surprise}, \emph{trust},
and \emph{anticipation}, it appears somewhat less obvious which emojis one would normally use to convey such emotions. Instead, we observe that individual annotators sometimes provided high rating scores based on idiosyncratic associations. One rater, for instance, associated a gemstone with a high degree of anticipation, while the others did not.
It is important to be aware of these varying correlation scores and compute separate correlation scores per emotion when evaluating emotion prediction models on this data. In Figure \ref{fig:pearson-fluctuation}, we visualize the emotion-specific agreement for different individual raters.

\paragraph{Emoji-Specific Agreement.} We also 
invoke Krippendorff's $\alpha$ as a measure of 
agreement between raters for each individual emoji along with its emotions. This allows us to understand to what extent the raters agree or disagree on the rating of a \emph{particular emoji--emotion pairing}.
Krippendorff's $\alpha$ scores range from 0 to 1, where $\alpha=0$ denotes no agreement and $\alpha=1$ represents the highest level of agreement among all users. 
Table \ref{tab:score-with-alpha} shows a few examples of emojis with specific emotions and their associated ratings, including the Krippendorff $\alpha$ value and standard deviation. The original intensity scores range from 0 to 4, but are rescaled to $[0,1]$.
We include examples with high agreement as well as ones with low agreement.

\paragraph{Distribution and Examples.} 
In Table \ref{tab:emoji-bucket}, we report the distribution of scores for different emotions. As one might reasonably expect, the lowest-intensity bucket is the largest for each considered emotion. Overall, fairly few emojis are strongly associated with \emph{anger}, \emph{disgust}, \emph{fear}, \emph{sadness}, or \emph{surprise}. For \emph{disgust}, there is no emoji in the highest-intensity bucket, although some emojis have a moderate intensity level. There are numerous emojis associated with \emph{anticipation}. The most atypical distribution is observed for \emph{joy}, as there appear to be a wide range of objects and concepts that spark joy, in addition to the emojis that directly express joy. 

Finally, in Table \ref{tab:emoji-group}, we list the top-ranked emotion-bearing emojis for each of the 8 considered emotions. 
Indeed, for many emotions, we encounter some of the most prototypically expected emojis, especially facial expression ones. Note that in some cases, common use diverges from the original Unicode definitions of the emojis, as for instance for the ``Persevering face'' \includegraphics[height=\fontcharht\font`\H]{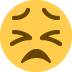} emoji, which is also associated with \emph{disgust} rather than just with perseverance.

\input{tab-emoji-bucket.tex}

%% file: tab-score-details.tex
\begin{table*}
\centering
\resizebox{\linewidth}{!}{
\begin{tabular}{|l|c|c|c|r|r|r|c|}
\hline
\textbf{Emoji} & \textbf{Name} & \textbf{Emotion} & \textbf{\makecell{All Ratings}} & \textbf{\makecell{Emotion\\Score}} & \textbf{K.\ $\mathbf{\alpha}$}  & \textbf{SD $\mathbf{\sigma}$}
\\ \hline

U+1F621	\includegraphics[height=\fontcharht\font`\H]{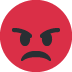} &	Pouting Face	&	Anger &	All 1.00s	& 1.00 &	\textcolor{mygreen}{0.61}	&	\textcolor{mygreen}{0.00} 
\\ \hline

U+1F60A \includegraphics[height=\fontcharht\font`\H]{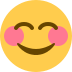} & \makecell{Smiling face with\\smiling eyes} & Joy & 6x1.00 and 3x0.75  & 0.92 &  \textcolor{mygreen}{0.68}	& \textcolor{mygreen}{0.12}
\\ \hline

U+1F62D \includegraphics[height=\fontcharht\font`\H]{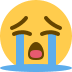} & \makecell{Loudly crying face} & Sadness & All 1.00s  & 1.00 &  \textcolor{mygreen}{0.48}	& \textcolor{mygreen}{0.00}
\\ \hline

U+1F633	\includegraphics[height=\fontcharht\font`\H]{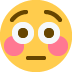} &	Flushed Face	&	Fear &	2x0.00, 2x0.25, 1x0.50, 2x0.75, 2x1.00  &	0.50	&	\textcolor{red}{0.12}	&	\textcolor{red}{0.37} 
\\ \hline

U+1F449 \includegraphics[height=\fontcharht\font`\H]{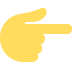} & \makecell{Backhand Index\\Pointing Right} & Anticipation & 6x0.00, 1x0.5, 1x0.75, 1x1.00 & 0.25 & \textcolor{red}{0.11}	& \textcolor{red}{0.37}
\\ \hline

\end{tabular}
}
\caption{Examples emoji emotion ratings along with Krippendorff's $\alpha$ and Standard Deviation $\sigma$.  
}
\label{tab:score-with-alpha}
\end{table*}

%% file: tab-emoji-group.tex
\begin{table}
\centering
\begin{tabular}{|c|l|l|}

\hline
\textbf{Emotion} & \textbf{Score} & \textbf{Emoji} \\ \hline
   & 1.00 & \includegraphics[height=\fontcharht\font`\H]{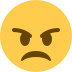} Angry face \\ %
Anger    & 1.00 & \includegraphics[height=\fontcharht\font`\H]{twitter-emoji-v2/1f621.png} Pouting face \\ %
    & 0.75 & \includegraphics[height=\fontcharht\font`\H]{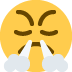} Face with steam\\& & ~~~~from nose \\ \hline

   & 0.97 & \includegraphics[height=\fontcharht\font`\H]{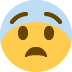} Fearful face \\ %
Fear    & 1.00 & \includegraphics[height=\fontcharht\font`\H]{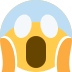} Face screaming \\& & ~~~~in fear \\ %
    & 0.90 & \includegraphics[height=\fontcharht\font`\H]{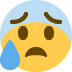} Anxious face \\& & ~~~~with fear \\ \hline

   & 1.00 & \includegraphics[height=\fontcharht\font`\H]{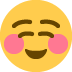} Smiling face \\
   Joy    & 0.94 & \includegraphics[height=\fontcharht\font`\H]{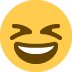} Grinning squinting\\& & ~~~~face \\ %
   & 0.94 & \includegraphics[height=\fontcharht\font`\H]{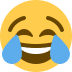} Face with tears \\& & ~~~~of joy \\ %
     \hline
    
   & 1.00 & \includegraphics[height=\fontcharht\font`\H]{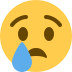} Crying face \\ %
Sadness & 1.00 & \includegraphics[height=\fontcharht\font`\H]{twitter-emoji-v2/1f62d.png} Loudly crying face \\ %
    & 0.94 & \includegraphics[height=\fontcharht\font`\H]{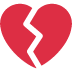} Broken heart \\ \hline

   & 0.81 & \includegraphics[height=\fontcharht\font`\H]{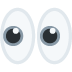} Eyes \\ %
Anticipation & 0.64 & \includegraphics[height=\fontcharht\font`\H]{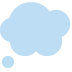} Thought balloon \\ %
    & 0.58 & \includegraphics[height=\fontcharht\font`\H]{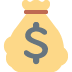} Money bag \\ \hline

   & 0.67 & \includegraphics[height=\fontcharht\font`\H]{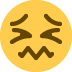} Confounded face \\ %
Disgust & 0.67 & \includegraphics[height=\fontcharht\font`\H]{twitter-emoji-v2/1f623.png} Persevering face \\ %
    & 0.67 & \includegraphics[height=\fontcharht\font`\H]{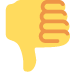} Thumbs down sign \\ \hline

   & 0.89 & \includegraphics[height=\fontcharht\font`\H]{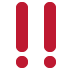} Double exclamation\\& & ~~~~mark \\ %
Surprise & 0.81 & \includegraphics[height=\fontcharht\font`\H]{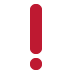} Exclamation mark \\ %
    & 0.69 & \includegraphics[height=\fontcharht\font`\H]{twitter-emoji-v2/1f631.png} Face screaming \\& & ~~~~in fear \\ \hline
    
   & 0.83 & \includegraphics[height=\fontcharht\font`\H]{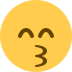} Kissing face with\\& & ~~~~smiling eyes \\ %
Trust & 0.75 & \includegraphics[height=\fontcharht\font`\H]{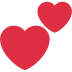} Two hearts \\ %
    & 0.72 & \includegraphics[height=\fontcharht\font`\H]{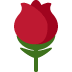} Rose \\ \hline
    
\end{tabular}
\caption{A glimpse of top emotion-intensive emojis for all eight emotions derived from the annotation experiment}
\label{tab:emoji-group}
\end{table}

%% file: tab-emoji-bucket.tex
\begin{table*}
\centering
\begin{tabular}{l r r r r r r r r}
\hline
\textbf{Groups} & \textbf{Anger} & \textbf{Anticipation} & \textbf{Disgust} & \textbf{~~~Fear} & \textbf{~~~Joy} & \textbf{Sadness}  & \textbf{Surprise} & \textbf{Trust} \\ \hline

B4 ($\geq$ 0.75)	&   3	&   1	&   0	&   3	&   \textbf{23}	&   6	&   2	&   2 \\ 
B3 ($\geq$ 0.50)    &   3	&   5	&   14	&   8	&   \textbf{24}	&   8	&   5	&   \textbf{24} \\
B2 ($\geq$ 0.25)    &   19	&   \textbf{86}	&   20	&   18	&   35	&   13	&   33	&   38 \\
B1 ($\geq$ 0.00)    &   \textbf{125}	&   58	&   116	&   121	&   68	&   123	&   110	&   86 \\ \hline

\end{tabular}
\caption{The distribution of 150 target emojis across four buckets B1, B2, B3, and B4 with respect to their gold intensity score for all 8 emotions. The bold score represents which emotion gets the highest number of emojis in the respective bucket}
\label{tab:emoji-bucket}
\end{table*}

%% file: sec-prediction.tex
\section{Emotion Scoring Experiments}
\label{sec:emotion-prediction-exp}
Given our manually collected data for 150 emojis, we next consider to what extent simple unsupervised methods can be used to reproduce such associations automatically in a data-driven manner. The data compiled in Section \ref{sec:annotation-task}, specifically the mean ratings for emoji--emotion pairs, serve as the ground truth.

We present several methods to predict emotion scores for emojis. 
In prediction, we not only rely on an existing English word--emotion lexicon but also make use of distributional similarity to support the prediction task. The latter is based on an emoji-centric corpus and resulting word vectors described in the following.

\subsection{Data}\label{sec:experiment:corpus}

In order to infer the correlation of emojis with major emotion-bearing words and vice versa, we created a web crawl of tweets collected specifically to provide emoji statistics by 
seeking out tweets containing at least one emoji.
We consider a set of 620 most frequently used emojis from \newcite{novak2015} and from Emoji Tracker%
. For each emoji, we then retrieved an equal number of tweets, labeled as being in English. In total, we obtained a set of 20.8 million tweets over a span of one year.

Subsequently, we train a simple word2vec Skip-Gram with Negative Sampling \cite{Mikolov-2013} model on this corpus. Because this corpus contains many occurrences of emojis, the resulting word vector representations include vectors for emojis, and we are able to compute the cosine similarity between emojis and words.

In the following, we explain how this data comes into play while predicting emotion ratings for any emojis available in our corpus.

\input{tab-results.tex}

\subsection{Methods}

We consider several methods to predict emoji--emotion association scores.

\paragraph{Prediction from Binary Emotion Labels.}
In this first approach, we consult EmoLex \cite{Mohammad13}, an existing English language word--emotion lexicon that assigns words binary labels for the same eight emotions that we consider in our study. Thus, a word may be tagged as being associated with \emph{trust} or as not being associated with it.
Specifically, we consult EmoLex to find the subset of words $\mathcal{V}_a$ from the vocabulary $\mathcal{V}$ that are associated with emotion (affect) $a$. 
We then rank the top $k=5$ such words based on the word--emoji cosine similarities induced earlier, and finally compute an emoji $e$'s emotion score $\sigma(e,a)$ for affect $a$ by summing up the similarity scores for the top $k$ words:
\begin{equation}
    \sigma(e,a)=\sum\limits_{w \in \mathrm{topk}(e,\mathcal{V}_a)} \mathrm{sim}(\mathbf{v}_w,\mathbf{v}_e),
\end{equation}
where $\mathrm{topk}(e,\mathcal{V}_a)$ denotes the top-$k$ words  $w \in \mathcal{V}_a$ in terms of $\mathrm{sim}(\mathbf{v}_w,\mathbf{v}_e)$ scores,
$\mathbf{v}_w$ denotes the word vector embedding for a word $w$,
$\mathbf{v}_e$ denotes the word vector embedding for an emoji $e$,
and $\mathrm{sim}(\cdot,\cdot)$ denotes the cosine similarity between vectors.

\paragraph{Prediction from Intensity Scores.}
Next, we consider emotion lexicons that, unlike EmoLex, provide real-valued emotion scores for English words.
In this case, the emotion intensity scores of words directly figure into the predicted scores. We first consult the lexicon to find all words $\mathcal{V}_a$ for which the lexicon provides any emotion intensity score at all for affect $a$.
We then identify the top $k$ words in terms of the word--emoji cosine similarity scores, as earlier.
Then, however, our predicted score $\sigma(e,a)$ is the
arithmetic mean of emotion scores of the top $k$ words.
Specifically,
\begin{equation}
    \sigma(e,a)=\frac{1}{k} \sum\limits_{w \in \mathrm{topk}(e,\mathcal{V}_a)} \tau(w,a),
\end{equation}
where $\tau(w,e)$ denotes the emotion intensity score provided by the lexicon and the remaining variables are defined as earlier.

In our experiments with this approach, we consider two separate word--emotion lexicons: the NRC Word Affect Intensity lexicon \cite{Saif-LREC18} and DepecheMood++ \cite{araque2018depechemood++}. 
The latter has a different emotion inventory than the  \newcite{Plutchik1980} emotion labels that we rely upon.
Thus, we apply the following mapping: \emph{angry} / \emph{annoyed} $\mapsto$ \emph{anger} (both reported separately), \emph{afraid} $\mapsto$ \emph{fear}, \emph{happy} $\mapsto$ \emph{joy}, \emph{sad} $\mapsto$ \emph{sadness}, and \emph{amused} $\mapsto$ \emph{surprise}. 

\paragraph{Prediction from Intensity Scores (Frequency-based).} We also consider a variant of the above formula, where $\mathrm{topk}(e,\mathcal{V}_a)$ considers not a ranking in terms of word2vec cosine similarities, but instead a ranking of words in terms of their co-occurrence frequency with the emoji $e$ in our Twitter corpus.

\subsection{Results}

Table \ref{tab:results} provides the Pearson Correlation score between the mean human-annotated emotion ratings and our predicted scores using the aforementioned methods, for a range of different parameter settings.

Using EmoLex with our binary emotion label scores, we observe a very low correlation for \emph{joy} and \emph{trust} and even negative correlation for \emph{anticipation} and \emph{surprise}. This appears to stem from the fact that binary emotion labels do not convey sufficient information for a more accurate prediction. The EmoLex lexicon merely signals whether or not it considers a word as being associated with an emotion.

With the NRC Word Affect Intensity lexicon \cite{Saif-LREC18}, we are able to obtain substantially higher correlations for a range of different settings of top-$k$ words, and with cosine similarity as well as with cooccurrence frequency rankings.
However, the NRC Word Affect Intensity lexicon cannot be applied to predict all emotion categories, as it provides emotion intensity scores only for \emph{anger}, \emph{fear}, \emph{joy}, and \emph{sadness}.

DepecheMood++ is as well unable to cover all emotions in our data. Since the lexicon also provides the frequency of the tokens observed in their collection, we report results for different frequency thresholds $F$ to refine the set of emotion-intensive words. 
While a few results are fairly good, overall the automatic data-driven induction process used to obtain DepecheMood++ does not yield as good results as the high-quality scores compiled in the NRC Affect Intensity lexicon. Interestingly, using the \emph{annoyed} label in DepecheMood++ rather than its \emph{angry} label, we obtain a higher correlation for \emph{anger}.

Overall, we find that we are able to obtain a high correlation with human ratings for those emotions for which we have accurate word emotion associations. Thus, we apply our models to predict scores for a larger set of 620 emojis. We will make this data publicly available as well.

%% file: tab-results.tex
\begin{table*}
\centering
\resizebox{\linewidth}{!}{
\begin{tabular}{|c|l|r|r|r|r|r|r|r|r|r|}
\hline
\textbf{Source} & \textbf{Settings} & \textbf{Anger} & \textbf{Anticipation} & \textbf{Disgust} & \textbf{~~~Fear} & \textbf{~~~~Joy} & \textbf{Sadness}  & \textbf{Surprise} & \textbf{Trust} & \textbf{Average} \\ \hline

EmoLex & $k=$5 & 0.62	& {-0.03}	    & \textbf{0.81}	&   0.50	& 0.19	& 0.57	 & -0.27	& \textbf{-0.04}	& 0.29 \\ \hline

\multirow{5}{*}{\makecell{Affect\\Intensity\\(Sim.\ Based)}} & $k=$10  & 0.35  &	N/A   & N/A     & 0.45	& 0.61	& 0.50 & N/A & N/A & 0.48 \\ 
& $k=$25 &0.43	& N/A & N/A &0.63	&0.63	&0.65 & N/A & N/A & 0.59 \\  
& $k=$100 & 0.57	& N/A	    & N/A	& {0.71}	& \textbf{0.71}	& {0.74}	 & N/A	& N/A	& {0.68} \\
& $k=$150 & 0.59	& N/A & N/A & \textbf{0.72}	&0.70	&0.72 & N/A &N/A & {0.68} \\
& $k=$300 & 0.61	& N/A & N/A & \textbf{0.72}	& \textbf{0.71}   &0.68   & N/A & N/A & {0.68} \\ \hline

\multirow{4}{*}{\makecell{Affect\\Intensity\\(Frequency Based)}} & $k=$25 &0.51	& N/A & N/A &0.61	&0.70	&0.74	& N/A & N/A & 0.64 \\
& $k=$50 & {0.63}	& N/A	    & N/A	&   0.57	& 0.69	& {0.71}	 & N/A	& N/A	& {0.65} \\
& $k=$100 & {0.69}	& N/A	    & N/A	&   0.66	& 0.67	& {0.75}	 & N/A	& N/A	& {0.69} \\
& $k=$150 & \textbf{0.70}	& N/A & N/A &0.68	&0.67	&0.75 & N/A & N/A & \textbf{0.70} \\	\hline

\multirow{12}{*}{\makecell{Depeche\\Mood++\\(Sim.\ Based)}} 
&	$k=$50,$F=$0	&	0.45/0.35	&	N/A	&	N/A	&	0.26	&	0.45	&	0.59	&	{0.08}	&	N/A	&	0.36 \\
&	$k=$50,$F=$25	&	0.49/0.45	&	N/A	&	N/A	&	0.31	&	0.49	&	0.61	&	0.09	&	N/A	&	0.41 \\
&	$k=$50,$F=$50	&	0.52/0.39	&	N/A	&	N/A	&	0.29	&	0.46	&	0.65	&	0.04	&	N/A	&	0.39 \\
&	$k=$100,$F=$0	&	0.41/0.34	&	N/A	&	N/A	&	0.38	&	0.46	&	0.70	&	0.08	&	N/A	&	0.40 \\
&	$k=$100,$F=$25	&	0.50/0.44	&	N/A	&	N/A	&	0.40	&	0.51	&	0.71	&	0.06	&	N/A	&	0.44 \\
&	$k=$100,$F=$50	&	0.53/0.43	&	N/A	&	N/A	&	0.38	&	0.53	&	0.73	&	0.06	&	N/A	&	0.44 \\
&	$k=$150,$F=$0	&	0.42/0.38	&	N/A	&	N/A	&	0.31	&	0.45	&	0.73    &	0.08	&	N/A	&	0.40 \\
&	$k=$150,$F=$25	&	0.51/0.43	&	N/A	&	N/A	&	0.37	&	0.44    &	0.75	&	0.08	&	N/A	&	0.43 \\
&	$k=$150,$F=$50	&	0.53/0.43	&	N/A	&	N/A	&	0.35	&	0.44    &	0.76	&	0.06	&	N/A	&	0.43 \\
&	$k=$200,$F=$0	&	0.47/0.39	&	N/A	&	N/A	&	0.27	&	0.40	&	0.74	&	0.11	&	N/A	&	0.40 \\
&	$k=$200,$F=$25	&	0.55/0.42	&	N/A	&	N/A	&	0.36	&	0.39	&	0.77	&	\textbf{0.12}	&	N/A	&	0.44 \\
&	$k=$200,$F=$50	&	0.58/0.43	&	N/A	&	N/A	&	0.37	&	0.41	&	\textbf{0.78}	&	0.10	&	N/A	&	0.44 \\ \hline

\end{tabular}
}
\caption{Pearson Correlation scores for all considered prediction methods. For DepecheMood++, anger column represents \emph{angry}/\emph{annoyed} and F represents the minimum frequency threshold. Bolded scores represent the highest correlation observed for the emotion in the respective column.}
\label{tab:results}
\end{table*}

%% file: sec-conclusion.tex
\section{Conclusion}
\label{sec:conclusion}
The desire to express an emotion is one of the factors that has driven the tremendous proliferation of emojis in interpersonal communication. In this work, we shed light on this connection by compiling a dataset that quantifies people's reported association between emojis and emotion. 
From each of 9 human raters, we solicit 1,200 ratings covering a set of 150 emojis with regard to 8 core emotions from \newcite{Plutchik1980}'s Wheel of Emotions.

This constitutes the first resource of this kind, which we thoroughly analyze and will make freely available to enable further research.

In terms of future work, one important avenue will be to  assess to what extent there may be cultural differences in these associations. A previous study found that the meaning of an emoji remains relatively stable across different languages and media \cite{Barbieri:2016:CEE:2964284.2967278}. In part, this may stem from the language-independent visual nature of emojis.
However, different concepts may have different associations in different cultures, which merits further study. Similarly, one could study variation with respect to age and other variables.

Finally, in our work, we rely on our annotated data to study how well we can automatically estimate emotional association ratings for a given emoji, considering a series of different baseline methods and resources. Our findings suggest that data-driven methods can fare quite well at this if combined with high-quality affective intensity information at the lexical level.
The success of our methods suggest that we are able to 
predict high-quality emotion scores for a much larger set of emojis. This opens up further research avenues on possible downstream applications exploiting this knowledge, such as consumer sentiment and emotion analytics, context-sensitive emoji recommendation, and computational social science.